\documentclass[10pt,twocolumn,letterpaper]{article}

\usepackage[pagenumbers]{wacv} %

\usepackage{booktabs}
\usepackage{colortbl}
\usepackage{tabularx}
\definecolor{oursrow}{gray}{0.90}
\definecolor{ctrlred}{RGB}{190,30,30}
\definecolor{ctrlblue}{RGB}{30,70,190}
\definecolor{redrow}{RGB}{252,226,226}
\definecolor{bluerow}{RGB}{224,234,252}
\definecolor{ctrlpurple}{RGB}{110,40,170}
\definecolor{purplerow}{RGB}{236,226,252}
\usepackage{subcaption}

\newcommand{\method}{Region-VAD\xspace}

\newcommand{\circled}[1]{\raisebox{0.4pt}{\textcircled{\raisebox{-0.6pt}{\scriptsize #1}}}}
\newcommand{\tiouLM}{TIoU\textsubscript{\cite{liu2019background}}}
\newcommand{\tiouSST}{TIoU\textsubscript{\cite{abdulaziz2026sst}}}
\usepackage{pifont}
\newcommand{\cmark}{\ding{51}}
\newcommand{\xmark}{\textcolor{gray}{\ding{55}}}

\definecolor{wacvblue}{rgb}{0.21,0.49,0.74}
\usepackage[pagebackref,breaklinks,colorlinks,allcolors=wacvblue]{hyperref}
\makeatletter
\g@addto@macro\normalsize{\setlength\abovedisplayskip{3pt plus 1pt minus 1pt}\setlength\belowdisplayskip{3pt plus 1pt minus 1pt}\setlength\abovedisplayshortskip{2pt}\setlength\belowdisplayshortskip{3pt}}
\renewcommand\paragraph{\@startsection{paragraph}{4}{\z@}{1pt plus 1pt minus 1pt}{-1em}{\normalfont\normalsize\bfseries}}
\newcommand{\keepwithnext}[1]{\par\vskip\z@\@plus #1\penalty-100\vskip\z@\@plus-#1\relax}
\def\wacvsection{\@startsection {section}{1}{\z@}{-6pt plus -2pt minus -2pt}{3pt} {\large\bf}}
\def\wacvsubsection{\@startsection {subsection}{2}{\z@}{-5pt plus -2pt minus -2pt}{2.5pt} {\elvbf}}
\makeatother

\def\wacvPaperID{20} %
\def\confName{WACV}
\def\confYear{2027}

\title{Region-VAD: Video Anomaly Localization and Detection via Region Selection}

\author{Giacomo D'Amicantonio\\
Eindhoven University of Technology\\
{\tt\small g.d.amicantonio@tue.nl}
}

\begin{document}
\makeatletter
\twocolumn[{
  \@maketitle
  \vspace{-14pt}
  \begin{center}
    \includegraphics[width=\textwidth]{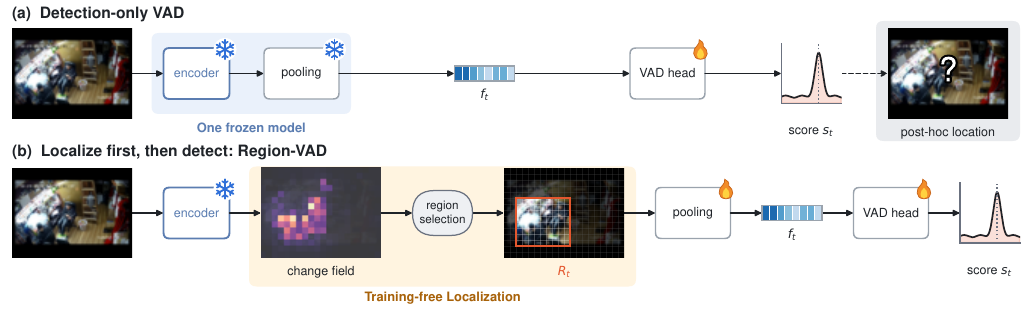}
    \vspace{-6pt}
    \captionof{figure}{\textbf{Localize to detect.} (a) Detection-only WS-VAD scores the pooled frozen features of the full frame with a trained head and, when it gives a location, derives it afterwards from the score (grey). (b) \method first selects a region $R_t$ from a training-free field over the frozen token grid (yellow) and then pools and scores only the tokens inside $R_t$. Flames mark trained components, snowflakes frozen ones.}
    \label{fig:teaser}
  \end{center}
  \vspace{0pt}
}]
\makeatother
\begin{abstract}
Anomalies are local events, yet weakly supervised video anomaly detection scores them globally. Most methods pool an entire frame into a single vector, rank frames by that vector, and derive a location afterwards as a saliency map of the score. Full-frame pooling mixes the anomaly with its background, so a detector can reach competitive benchmark scores from scene context without being able to show where in the scene the anomaly occurs. We introduce \method, which first selects a region and then detects the anomaly from that region alone. \method selects an explicit rectangle over the tokens of a frozen foundation model and scores anomalousness only from the tokens inside the box, through attention pooling followed by a standard temporal head. The rectangle is not learned, it is the box in which the frozen tokens change most between consecutive segments, in all four feature projections of the encoder, and it is found by exhaustive search over the rectangles of the token grid. Only the pooling and the temporal head are trained, from video-level labels without boxes, masks or frame-level annotations, and no trained component reads a token outside the selected region. Experiments on five datasets, with one localizer configuration shared by all datasets and backbones, show that \method achieves state-of-the-art anomaly localization on UCF-Crime and MSAD at competitive detection performance.
\end{abstract}
\vspace{-6pt}
    
\section{Introduction}
\label{sec:intro}

Weakly supervised video anomaly detection (WS-VAD) aims to identify the frames of a video in which an anomalous event occurs, given only a video-level label for training~\cite{sultani2018real}. Most methods pool each clip into a single vector and rank the clips, and when a location is needed they derive it afterwards as a map of the clip score~\cite{sultani2018real,liu2019background} (Fig.~\ref{fig:teaser}a). The location matters for two reasons. An operator needs to know \emph{where} the event takes place, and a detector that shows what it attended to can be checked for whether it learned the scene or the event~\cite{liu2019background,abdulaziz2026sst}. We address both detection and localization from video-level labels alone.

Spatial localization has been added to the multiple-instance learning (MIL) pipeline~\cite{sultani2018real,tian2021rtfm,zhou2023urdmu} through tube proposals~\cite{wu2021wsstad,li2022scale}, vision--language alignment maps~\cite{wu2024vadclip,wu2024stprompt}, top-scoring patches~\cite{karim2026patch} and learned token sparsification~\cite{abdulaziz2026sst}. In all of them, the spatial output does not constrain the temporal score. The detector remains free to exploit the entire frame, and since anomalous and normal videos can often be separated by scene context alone, it can reach competitive accuracy while attending to the background~\cite{liu2019background}. Moreover, the predicted box is rarely verified as the IoU metric rewards both \emph{where} a model looked and \emph{how large} a box it drew, yet no published result is compared with a box of the same size at the frame centre. On XD-Violence, even a constant full-frame box is competitive. Finally, most localizers add a trained component and a dependency (proposals, detectors, prompts or a second encoder), so their localization cannot be separated from what they were trained with.

\method reverses this order (Fig.~\ref{fig:teaser}b). It first selects a region without training and then detects the anomaly from that region. When the token of a patch changes between consecutive segments, something has moved or appeared there. \method rests on two common hypotheses in the field: the region in which the frozen tokens change is the region of the event and can be found without training, and the tokens inside it carry the evidence needed to detect the event, so a detector that reads only them outperforms one that reads the whole frame. Therefore, \method leverages C-RADIOv4~\cite{ranzinger2024radio,radiov4}, which describes each frame in four projections: its own backbone's features and those of its DINOv3, SAM~3, and SigLIP~2-aligned heads~\cite{simeoni2025dinov3,sam3,tschannen2025siglip2}. A change seen in all four is unlikely to be spurious, and highlights the region in which events are happening in the frame. \method standardizes the change map of each projection, keeps in every patch only the change that all four report, smooths the result over the neighboring segments, and selects by exhaustive search the rectangle with the largest centered sum of the resulting field. A standard temporal head then scores an attention-pooled vector of the tokens inside the rectangle. Only the pooling and the head are trained from video-level labels alone, while the localizer has no learnable parameters. 

Our contributions are summarized as follows:
\begin{itemize}[leftmargin=*,itemsep=1pt,topsep=2pt]
  \item We propose \method, a localize-then-detect architecture for WS-VAD that combines a training-free region localizer on a frozen token grid, attention pooling confined to it and a standard temporal head, trained from video-level labels only;
  \item We introduce three controls for spatial localization against which we report every result: the size-matched centered box, the full-frame box, and a fixed box at the center of the frame;
  \item We achieve state-of-the-art localization on UCF-Crime and MSAD, and show that a standard temporal head detects more accurately from the tokens inside the box than from the full frame on all evaluated benchmarks.
\end{itemize}

\vspace{-6pt}
\section{Related Work}
\label{sec:related}
\begin{figure*}[t]
  \centering
  \includegraphics[width=\textwidth]{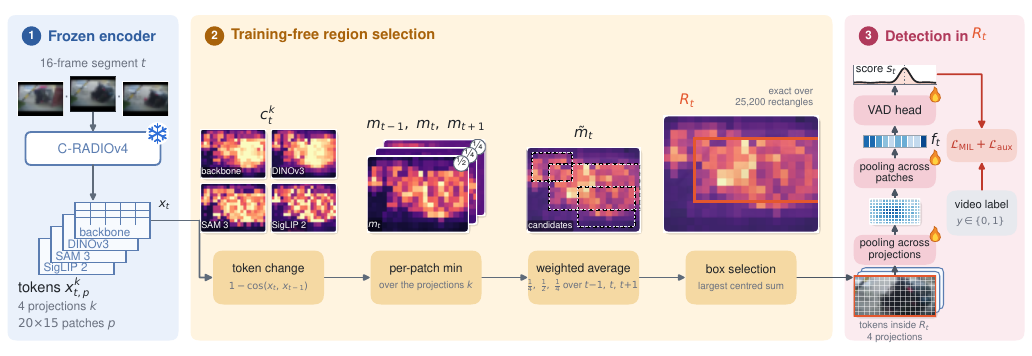}
  \caption{\textbf{Overview of \method.} \circled{1} A frozen C-RADIOv4 encoder maps the middle frame of each 16-frame segment to a $20\times15$ token grid in four projections. \circled{2} Region selection has no learnable parameter. The standardized token change from segment $t-1$ to $t$ gives one map per projection. Their per-patch minimum, the consensus $m_t$, is averaged over $t-1$, $t$, $t+1$ into $\tilde m_t$, and $R_t$ is the rectangle with the largest centred sum of $\tilde m_t$. \circled{3} The tokens inside $R_t$ are pooled by two trained attention stages, across the four projections into one map of $d$-dimensional tokens and then across the patches into $f_t\in\mathbb{R}^d$, which a VAD head scores. The video label supervises only the blocks marked with a flame. Fields and boxes are outputs of the method on UCF-Crime RoadAccidents121.}
  \label{fig:arch}
  \vspace{-16pt}
\end{figure*}

\paragraph{Weakly supervised video anomaly detection.}
The dominant paradigm was introduced with UCF-Crime~\cite{sultani2018real}, where clips embedded by a frozen backbone~\cite{tran2015c3d,carreira2017i3d} are scored by a network trained with a multiple-instance ranking loss. Later work improved the scorer~\cite{wu2024review,majhi2025pivad,damicantonio2025gsmoe} with pseudo-labels~\cite{zhong2019gcn,feng2021mist,li2022msl,sun2023lstc}, feature-magnitude and normalcy objectives~\cite{tian2021rtfm,zaheer2020claws,chen2023mgfn}, memory units~\cite{zhou2023urdmu} and vision-language backbones~\cite{wu2024vadclip,yang2024tpwng,pu2024pel,wu2024ovvad}. The benchmarks expanded to ShanghaiTech~\cite{liu2018future,zhong2019gcn}, XD-Violence~\cite{wu2020xdv}, UBnormal~\cite{acsintoae2022ubnormal} and MSAD~\cite{zhu2024msad}, and the metrics have been re-examined~\cite{liu2025metrics}. The tendency of the scorer to exploit context has been countered in the temporal feature space, with context-invariant features~\cite{lv2023umil}, context-motion relations~\cite{cho2023lookaround} or separate human and scene streams~\cite{majhi2023humanscene}. Training-free variants replace the scorer with a captioner and a language model~\cite{zanella2024lavad}. All of these methods score a feature of the entire frame.

\paragraph{Spatial anomaly localization.}
In the one-class setting, object-centric methods crop every object found by a pretrained detector and score the crops~\cite{ionescu2019objectcentric,georgescu2021backgroundagnostic,georgescu2021multitask,wang2022jigsaw}. They localize by construction~\cite{ramachandra2020streetscene}, but only for the classes of the detector, so an anomaly of another class, or one without a distinct object, is never cropped. An open-vocabulary segmenter such as SAM~3~\cite{sam3} could widen this set at the cost of a further large model. \method also localizes before scoring, but takes its region from the change of the encoder's own tokens between consecutive segments, without a detector or training, so the region is not tied to object classes. It also learns from video-level labels instead of normal videos alone. Under weak supervision, boxes were first annotated on UCF-Crime to show that training on the local region improves detection~\cite{landi2019locality}, and the annotation of its test set~\cite{liu2019background} introduced the per-frame box IoU and documented that detectors exploit the background. Later localizers obtain the box from tube proposals ranked under the MIL objective~\cite{wu2021wsstad,li2022scale}, from the alignment of local features with text embeddings~\cite{wu2024stprompt,wu2024vadclip}, or from a $7\times7$ patch grid with a proximity-aware top-$k$~\cite{karim2026patch}. In each case the region is derived from a clip score computed on the full frame, and none is compared with a box of fixed size. Closest to our motivation, SST-WSVADL~\cite{abdulaziz2026sst} learns to sparsify anomaly-relevant tokens under a motion-aware regularizer, releases dense boxes for UCF-Crime, XD-Violence and MSAD, and proposes the localization output as an audit of scene bias. We adopt its annotations and its auditing view, but our trained detector reads only the tokens inside the region.

\paragraph{Region selection over frozen tokens.}
Selecting a region and pooling inside it can be seen as multiple-instance learning with spatial instances~\cite{ilse2018attention}. Class activation maps~\cite{zhou2016cam,selvaraju2017gradcam} instead derive the region from a classifier after it has scored the image. Training-free localization from frozen tokens exists for single images, from a seed patch~\cite{simeoni2021lost}, a normalized cut of the patch graph~\cite{wang2022tokencut,melas2022deepspectral,wang2023cutler} or the attention of the background~\cite{simeoni2023found}, and motion has been used as an unsupervised cue for segmentation~\cite{choudhury2022gwm}. Our field instead measures the change of the tokens over time. Our tokens come from frozen foundation encoders, DINOv3~\cite{oquab2023dinov2,simeoni2025dinov3}, SAM~\cite{kirillov2023sam,ravi2024sam2,sam3} and SigLIP~2~\cite{tschannen2025siglip2}, distilled by C-RADIOv4~\cite{ranzinger2024radio,heinrich2025radiov25,radiov4} into one student that exposes a projection into each teacher's space. Motion has long served as a localization cue in anomaly detection, from flow-based models~\cite{mahadevan2010mdt,liu2018future} to the motion-aware regularizer of~\cite{abdulaziz2026sst}, and temporal differencing in feature space serves action recognition~\cite{wang2021tdn}. Our change field is computed from the encoder's own tokens in four projections at once, has no learned weight and selects the region itself, whereas~\cite{abdulaziz2026sst} uses motion to regularize a trained selector.

\vspace{-3pt}
\vspace{-4pt}
\section{Method}
\label{sec:method}

\method consists of three stages (Fig.~\ref{fig:arch}). A frozen encoder maps every segment of a video to a grid of tokens in four projections (Sec.~\ref{sec:prelim}). A training-free rule then selects one rectangle $R_t$ per segment from these tokens alone (Sec.~\ref{sec:select}). Finally, only the tokens inside $R_t$ are pooled and scored by a standard temporal head. This last stage is the only trained one and learns from video-level labels (Sec.~\ref{sec:detect}).

\subsection{Preliminaries: the frozen token grid}
\label{sec:prelim}
\begin{figure*}[t]
  \centering
  \includegraphics[width=\textwidth]{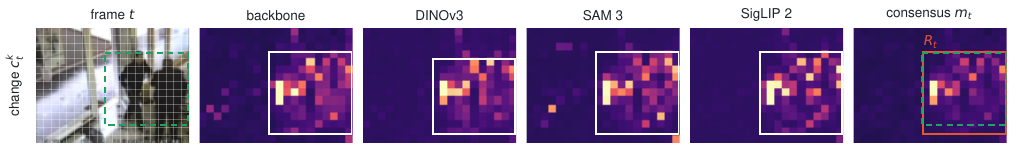}
  \caption{\textbf{Change maps of one frame.} UCF-Crime Fighting042, with the ground truth of~\cite{liu2019background} dashed in green. The change $c^k_t$ from segment $t-1$ to $t$ in the four projections of C-RADIOv4, standardized and on one colour scale, and their consensus $m_t$ (white: the rectangle selected from each projection's map alone; orange: $R_t$, selected from the temporal average $\tilde m_t$). Each projection marks patches that the others miss, and the consensus keeps the patches marked by all four.}
  \label{fig:views}
  \vspace{-16pt}
\end{figure*}

\paragraph{Encoder.}
We use C-RADIOv4~\cite{radiov4} in its SO400M variant, an agglomerative ViT with patch size 16~\cite{ranzinger2024radio,heinrich2025radiov25} distilled to reproduce the dense features of DINOv3~\cite{simeoni2025dinov3}, SAM~3~\cite{sam3} and SigLIP~2~\cite{tschannen2025siglip2} in a single forward pass. The encoder is frozen. A video is divided into segments of 16 frames, $t=1,\dots,T$, each represented by its middle frame, which is resized to $320\times240$ pixels without cropping. The encoder returns one 1152-dimensional token per $16\times16$ patch, a grid $P$ of $20\times15=300$ patches $p$.

\paragraph{Projection heads.}
C-RADIOv4 is released with the feature-projection heads of its distillation, which map the backbone tokens into the patch-feature space of each teacher. Applied to every encoded frame, they yield three further grids of the same layout: DINOv3 (4096-d), SAM~3 (1024-d) and SigLIP~2 (1536-d). The DINOv3 and SAM~3 heads are per-token perceptrons, whereas the SigLIP~2 head prepends two self-attention blocks, so its projection also depends on the rest of the frame. The heads are also frozen. We refer to the token of patch $p\in P$ in segment $t$ under projection $k\in\{1,2,3,4\}$ as $x^{k}_{t,p}\in\mathbb{R}^{d_k}$, with $k=1$ the backbone and $k=2,3,4$ the DINOv3, SAM~3 and SigLIP~2 heads respectively.

The four grids describe the same pixels in feature spaces shaped by different objectives and differ in granularity. DINOv3 and the backbone group the frame into a few smooth regions, while SAM~3 and SigLIP~2 vary almost token by token. They differ accordingly in how they measure change from one segment to the next (Fig.~\ref{fig:views}). The maps agree on the moving object but peak in different patches, and each reports changes that the others do not confirm.

\subsection{Training-free region selection}
\label{sec:select}
Where a token of the grid differs from the previous segment, something has moved or appeared, and such change is a cue for anomalous events that unfold over time. A video-level label does not say where to look, since every location in an anomalous video is equally consistent with it. We therefore compute the region from the frozen tokens with no learnable parameter (Fig.~\ref{fig:arch}, \circled{2}), so the localization of \method depends only on the tokens of the frozen encoder.

\paragraph{Token change.}
For every patch and projection, change is the cosine distance between a token of the current frame and the token of the same patch in the previous one,
\begin{equation}
  c^{k}_{t}(p)=1-\cos\!\left(x^{k}_{t,p},\,x^{k}_{t-1,p}\right),
  \label{eq:change}
\end{equation}
with the first segment compared to the second. We use the cosine because token norms vary across patches and projections independently of content, and a scale-free distance can be compared across them. A patch whose content and surroundings are unchanged has $c^{k}_{t}(p)\approx0$, and a patch in which an object moves or deforms changes more than the rest of the frame (Fig.~\ref{fig:views}). Since the four projections have different dynamic ranges, each map is standardized over the grid, $z(c)=(c-\mu)/\sigma$ with $\mu$ and $\sigma$ computed over $p\in P$. Standardization removes the uniform part of a change shared by the whole frame, such as camera motion or illumination.

\paragraph{Per-patch minimum.}
Since the four maps agree on the moving object (Fig.~\ref{fig:views}), we keep a change only where all four report it,
\begin{equation}
  m_t(p)=\min_{k}\; z\!\left(c^{k}_{t}\right)(p),
  \label{eq:consensus}
\end{equation}
the per-patch minimum of the standardized maps, the strictest consensus, which is high only in a patch that lies above the frame average in all four. For an encoder with a single projection, the minimum is that map itself and the rule is otherwise unchanged.

\paragraph{Temporal smoothing.}
An event spans several segments, whereas $m_t$ compares only two, and a change confined to a single pair of segments is more often a flicker than an event. We therefore average $m_t$ over the neighbouring segments,
\begin{equation}
  \tilde m_t=\tfrac14\,m_{t-1}+\tfrac12\,m_t+\tfrac14\,m_{t+1},
  \label{eq:smooth}
\end{equation}
with $\tilde m_1=(2m_1+m_2)/3$ and $\tilde m_T=(m_{T-1}+2m_T)/3$ at the ends of the video. Since $\tilde m_t$ uses $m_{t+1}$, $R_t$ is known only after the next segment is encoded, which delays it by 16 frames. The average favours a location that changes in consecutive segments over one that changes once.

\begin{table}[t]
  \centering
  \footnotesize
  \setlength{\tabcolsep}{2.5pt}
  \begin{tabularx}{\columnwidth}{@{}l*{4}{>{\centering\arraybackslash}X}|*{2}{>{\centering\arraybackslash}X}@{}}
    \toprule
    & \multicolumn{4}{c}{area (\% of the frame)} & \multicolumn{2}{c}{\shortstack{centre\\distance (px)}} \\
    \cmidrule(lr){2-5}\cmidrule(lr){6-7}
    & \multicolumn{2}{c}{annotated} & \multicolumn{2}{c}{$\boldsymbol{R_t}$ (ours)} & annot. & $\boldsymbol{R_t}$ \\
    \cmidrule(lr){2-3}\cmidrule(lr){4-5}\cmidrule(lr){6-6}\cmidrule(lr){7-7}
    Annotation set & mean & med. & mean & med. & mean & mean \\
    \midrule
    UCF-Crime~\cite{liu2019background} & 23.1 & 20.1 & 29.5 & 27.0 & 52.8 & 54.1 \\
    UCF-Crime~\cite{abdulaziz2026sst} & 28.4 & 23.5 & 30.6 & 28.0 & 58.8 & 55.3 \\
    MSAD~\cite{abdulaziz2026sst} & 28.0 & 19.4 & 27.2 & 25.7 & 61.8 & 61.1 \\
    XD-Violence~\cite{abdulaziz2026sst} & 69.1 & 76.2 & 47.2 & 48.0 & 22.0 & 36.6 \\
    UBnormal & 21.8 & 17.1 & 18.3 & 16.0 & 56.3 & 72.7 \\
    ShanghaiTech & 3.9 & 1.9 & 25.1 & 24.0 & 91.3 & 65.4 \\
    \bottomrule
  \end{tabularx}
  \caption{\textbf{Size and position of the boxes.} Mean and median area (\% of the frame) of the annotated boxes and of $R_t$ over the annotated anomalous test frames, and mean Euclidean distance (pixels of the $320\times240$ frame) from the box centre to the frame centre.}
  \label{tab:boxsize}
  \vspace{-12pt}
\end{table}

\paragraph{Box scoring and selection.}
The region is the rectangle of patches with the largest centred sum of the field,
\begin{equation}
  R_t=\arg\max_{R\in\mathcal{R}}\;\sum_{p\in R}\bigl(\tilde m_t(p)-\bar m_t\bigr),
  \label{eq:rect}
\end{equation}
where $\bar m_t$ is the mean of $\tilde m_t$ over the grid and $\mathcal{R}$ is the set of axis-aligned rectangles of the grid, from a single patch to the full frame. Centring fixes the sign of each patch's contribution, so that a patch raises the score of a rectangle if it changes more than the frame on average, $\tilde m_t(p)>\bar m_t$, and lowers it otherwise. $R_t$ is the rectangle with the largest excess of change over the frame's own level. The exhaustive search returns the maximum-sum subwindow of the centred field, as in efficient subwindow search~\cite{lampert2008ess}. The score of each of the $25{,}200$ rectangles of the grid is obtained from summed-area tables~\cite{crow1984sat} (\emph{candidates} in Fig.~\ref{fig:arch}) and the maximum is taken (\emph{box selection}), at a cost negligible next to the encoder. The rule has no tuned threshold, and the size of the box follows from the mean of the field, its only reference. We use no size prior, since one tuned to a dataset would carry into the next. The segment length and the three smoothing weights are fixed once for every dataset and encoder. A rectangle is selected in every segment, normal or anomalous. In a normal segment it encloses the patches that change most, and the head has to tell the two cases apart.

\subsection{Detection inside the rectangle}
\label{sec:detect}
The detection head reads only the tokens inside $R_t$, although these tokens carry information from the rest of the frame through the attention layers of the frozen encoder. Two properties of these tokens dictate how they are pooled. First, each patch carries four projections that describe it at different granularities (Sec.~\ref{sec:prelim}), and a first attention stage learns to weight the four per patch. Second, the rectangle is a coarse box that contains background patches next to the event, and a second stage weights the patches instead of averaging them. Both stages are attention probes~\cite{lee2019settransformer,zhai2023siglip}, a single learned query attending over a set (Fig.~\ref{fig:arch}, \circled{3}). For a segment $t$, write $x^{k}_{p}$ for the token $x^{k}_{t,p}$ of a patch $p\in R_t$.

\begin{table*}[t]
  \centering
  \footnotesize
  \renewcommand{\arraystretch}{0.84}
  \setlength{\tabcolsep}{2.2pt}
  \begin{tabularx}{\textwidth}{@{}ll*{11}{>{\centering\arraybackslash}X}@{}}
    \toprule
    & & \multicolumn{3}{c}{UCF-Crime} & \multicolumn{2}{c}{MSAD} & \multicolumn{2}{c}{XD-Violence} & \multicolumn{2}{c}{UBnormal} & \multicolumn{2}{c}{ShanghaiTech} \\
    \cmidrule(lr){3-5}\cmidrule(lr){6-7}\cmidrule(lr){8-9}\cmidrule(lr){10-11}\cmidrule(lr){12-13}
    Method & Backbone & \tiouLM{} & \tiouSST{} & AUC & \tiouSST{} & AUC & \tiouSST{} & AP & TIoU & AUC & TIoU & AUC \\
    \midrule
    \multicolumn{13}{@{}l}{\textsc{Controls}} \\
    \rowcolor{redrow}
    Full frame & C-RADIOv4 & 23.07 & 28.42 & 85.94 & 27.54 & 87.98 & \underline{68.98} & 71.44 & 21.65 & 63.52 & 3.85 & \underline{96.96} \\
    \rowcolor{bluerow}
    Centred box & C-RADIOv4 & 34.02 & 30.94 & 84.84 & 26.33 & 89.03 & 50.21 & 75.03 & 20.36 & 59.88 & 6.20 & 95.61 \\
    \rowcolor{purplerow}
    Fixed $8{\times}8$-patch box & C-RADIOv4 & 30.76 & 27.08 & 85.48 & 22.48 & 88.39 & 26.66 & 71.70 & 21.62 & 64.01 & 5.94 & 96.55 \\
    Frame-difference box & pixels & \textbf{45.96} & \underline{41.25} & 85.65 & \underline{40.57} & 89.55 & 44.76 & 76.83 & \textbf{48.71} & 67.45 & \textbf{12.92} & 94.90 \\
    \midrule
    Background-aware MIL~\cite{liu2019background} & I3D & 16.40 & -- & 82.00 & -- & -- & -- & -- & -- & -- & -- & -- \\
    Deep MIL~\cite{sultani2018real} & I3D & 16.82 & -- & 67.11 & -- & -- & 32.69$^{\dagger}$ & -- & 9.44$^{\dagger}$ & -- & 2.46 & -- \\
    WSSTAD~\cite{wu2021wsstad} & C3D & 8.98 & -- & 87.65 & -- & -- & -- & -- & -- & -- & -- & -- \\
    VadCLIP~\cite{wu2024vadclip} & CLIP & 22.05 & -- & 88.02 & -- & -- & -- & \underline{84.51} & -- & -- & 4.09 & -- \\
    STPrompt~\cite{wu2024stprompt} & CLIP & 23.90 & -- & 88.08 & -- & -- & -- & -- & -- & -- & 9.77 & -- \\
    Patch-MIL~\cite{karim2026patch} & UniFormer & 40.51 & -- & -- & -- & -- & 39.58$^{\dagger}$ & -- & 19.76$^{\dagger}$ & -- & \underline{12.56} & -- \\
    SST-WSVADL~\cite{abdulaziz2026sst} & I3D & -- & 27.28 & 86.50 & 34.97 & 89.00 & \textbf{72.50} & 82.39 & -- & -- & -- & -- \\
    SST-WSVADL~\cite{abdulaziz2026sst} & VideoMAE~v2 & 27.92 & 29.90 & \underline{88.51} & 35.02 & \textbf{94.17} & 63.09 & \textbf{86.00} & -- & -- & -- & -- \\
    \midrule
    \method, mean pooling & C-RADIOv4 & \underline{45.71} & \textbf{42.45} & 87.25 & \textbf{41.91} & 90.80 & 52.35 & 79.73 & \underline{44.68} & \underline{67.83} & 11.93 & 95.92 \\
    \rowcolor{oursrow}
    \textbf{\method (ours)} & C-RADIOv4 & \underline{45.71} & \textbf{42.45} & \textbf{88.93} & \textbf{41.91} & \underline{91.39} & 52.35 & 81.50 & \underline{44.68} & \textbf{68.17} & 11.93 & \textbf{97.13} \\
    \midrule
    \rowcolor{redrow}
    $\Delta$ Full frame & & \textcolor{ctrlred}{$+$22.6} & \textcolor{ctrlred}{$+$14.0} & \textcolor{ctrlred}{$+$3.0} & \textcolor{ctrlred}{$+$14.4} & \textcolor{ctrlred}{$+$3.4} & \textcolor{ctrlred}{$-$16.6} & \textcolor{ctrlred}{$+$10.1} & \textcolor{ctrlred}{$+$23.0} & \textcolor{ctrlred}{$+$4.7} & \textcolor{ctrlred}{$+$8.1} & \textcolor{ctrlred}{$+$0.2} \\
    \rowcolor{bluerow}
    $\Delta$ Centred box & & \textcolor{ctrlblue}{$+$11.7} & \textcolor{ctrlblue}{$+$11.5} & \textcolor{ctrlblue}{$+$4.1} & \textcolor{ctrlblue}{$+$15.6} & \textcolor{ctrlblue}{$+$2.4} & \textcolor{ctrlblue}{$+$2.1} & \textcolor{ctrlblue}{$+$6.5} & \textcolor{ctrlblue}{$+$24.3} & \textcolor{ctrlblue}{$+$8.3} & \textcolor{ctrlblue}{$+$5.7} & \textcolor{ctrlblue}{$+$1.5} \\
    \rowcolor{purplerow}
    $\Delta$ Fixed box & & \textcolor{ctrlpurple}{$+$15.0} & \textcolor{ctrlpurple}{$+$15.4} & \textcolor{ctrlpurple}{$+$3.5} & \textcolor{ctrlpurple}{$+$19.4} & \textcolor{ctrlpurple}{$+$3.0} & \textcolor{ctrlpurple}{$+$25.7} & \textcolor{ctrlpurple}{$+$9.8} & \textcolor{ctrlpurple}{$+$23.1} & \textcolor{ctrlpurple}{$+$4.2} & \textcolor{ctrlpurple}{$+$6.0} & \textcolor{ctrlpurple}{$+$0.6} \\
    $\Delta$ Frame-difference box & & $-$0.25 & $+$1.2 & $+$3.3 & $+$1.3 & $+$1.8 & $+$7.6 & $+$4.7 & $-$4.0 & $+$0.7 & $-$1.0 & $+$2.2 \\
    \bottomrule
  \end{tabularx}
  \caption{\textbf{State-of-the-art comparison on five benchmarks.} Localization and detection of the methods with a spatial output and of four controls. Three box controls test whether a box scores by its placement or by its size, and the frame-difference box applies the rule of Sec.~\ref{sec:select} to pixel differences. For detection, the same head scores the mean of each control's tokens (Sec.~\ref{sec:setup}). \tiouLM{}: boxes of~\cite{liu2019background} on every 16th frame; \tiouSST{}: dense boxes of~\cite{abdulaziz2026sst}; UBnormal and ShanghaiTech: official masks; $^{\dagger}$: unreleased keyframe boxes of~\cite{karim2026patch}, not comparable with ours. Numbers of published methods are those reported in~\cite{karim2026patch}, \cite{abdulaziz2026sst} and the respective papers. Bold and underline mark the \textbf{best} and \underline{second-best} per column, including the controls, and ties share the mark. Last four rows: gain of \method over the four controls.}
  \label{tab:main}
  \vspace{-10pt}
\end{table*}

\paragraph{Pooling.}
Each projection is first brought to a common dimension $d=512$ by its own layer normalization and linear layer, $h^{k}_{p}=W^{k}\,\mathrm{LN}^{k}(x^{k}_{p})\in\mathbb{R}^{d}$. Each probe $\mathrm{MAP}(\cdot)$ has $H=8$ heads followed by a residual MLP. The first probe pools the four projections of each patch, and a second probe with its own parameters pools the patches of the rectangle,
\begin{equation}
  g_p=\mathrm{MAP}\bigl(\{h^{k}_{p}\}_{k=1}^{4}\bigr),\qquad f_t=\mathrm{MAP}'\bigl(\{g_p\}_{p\in R_t}\bigr),
  \label{eq:pool}
\end{equation}
which gives one $d$-dimensional token $g_p$ per patch of $R_t$ and then one feature vector $f_t\in\mathbb{R}^{d}$ per segment.

\paragraph{Temporal head.}
The segment features are fed to the input layer of UR-DMU~\cite{zhou2023urdmu}, which keeps its published architecture, losses and loss weights and returns one anomaly score $s_t$ per segment. With the encoder frozen and the head unchanged, the region and the pooling are the only difference between \method and the full-frame pipeline of Fig.~\ref{fig:teaser}a. The two pooling stages and the head are trained jointly, from the video label $y\in\{0,1\}$ alone, with
\begin{equation}
  \mathcal{L}=\mathcal{L}_{\mathrm{MIL}}+\mathcal{L}_{\mathrm{aux}},
  \label{eq:loss}
\end{equation}
where $\mathcal{L}_{\mathrm{MIL}}$ is the binary cross-entropy between $y$ and the mean of the $n$ highest segment scores, $n=\lfloor T/16\rfloor+1$, and $\mathcal{L}_{\mathrm{aux}}$ collects the memory and KL terms of UR-DMU. The pooling has no loss of its own and learns which projections and patches serve the head only through the head's loss. $R_t$ is computed once from the frozen tokens and enters training as a fixed set of patches, so no gradient reaches it.

\begin{table}[t]
  \centering
  \footnotesize
  \renewcommand{\arraystretch}{0.88}
  \setlength{\tabcolsep}{1.5pt}
  \begin{tabularx}{\columnwidth}{@{}*{4}{>{\centering\arraybackslash}X}>{\hsize=1.12\hsize\centering\arraybackslash}X*{2}{>{\hsize=0.94\hsize\centering\arraybackslash}X}@{}}
    \toprule
    \multicolumn{4}{c}{token-change maps} & & \multicolumn{2}{c}{UCF-Crime} \\
    \cmidrule(lr){1-4}\cmidrule(lr){6-7}
    backbone & DINOv3 & SAM~3 & SigLIP~2 & smoothing & \tiouLM{} & \tiouSST{} \\
    \midrule
    \cmark & \xmark & \xmark & \xmark & \cmark & 42.75 & 40.71 \\
    \xmark & \cmark & \xmark & \xmark & \cmark & \underline{45.60} & \textbf{42.45} \\
    \xmark & \xmark & \cmark & \xmark & \cmark & 43.26 & 40.36 \\
    \xmark & \xmark & \xmark & \cmark & \cmark & 45.28 & \underline{42.34} \\
    \cmark & \cmark & \cmark & \cmark & \xmark & 44.24 & 40.20 \\
    \rowcolor{oursrow}
    \cmark & \cmark & \cmark & \cmark & \cmark & \textbf{45.71} & \textbf{42.45} \\
    \bottomrule
  \end{tabularx}
  \caption{\textbf{Ablation of the region selection on UCF-Crime.} TIoU of the rectangle selected by the rule of Sec.~\ref{sec:select} from one map or from their consensus, with or without the temporal smoothing. The grey row is \method{} (ours), and the controls are in Table~\ref{tab:main}. Bold/underline: best/second-best per column; ties share the mark.}
  \label{tab:stage1}
  \vspace{-14pt}
\end{table}

\begin{table}[t]
  \centering
  \footnotesize
  \renewcommand{\arraystretch}{0.88}
  \setlength{\tabcolsep}{2pt}
  \begin{tabularx}{\columnwidth}{@{}l*{6}{>{\centering\arraybackslash}X}@{}}
    \toprule
    & \multicolumn{3}{c}{UR-DMU~\cite{zhou2023urdmu}} & \multicolumn{3}{c}{MLP~\cite{sultani2018real}} \\
    \cmidrule(lr){2-4}\cmidrule(lr){5-7}
    Tokens & frame & mean & attn. & frame & mean & attn. \\
    \midrule
    backbone & 85.94 & 86.30 & 86.81 & \underline{86.35} & 86.23 & \underline{86.87} \\
    DINOv3 & 84.31 & \underline{86.83} & 85.47 & 84.77 & \underline{86.27} & 84.98 \\
    SAM~3 & 81.99 & 82.64 & 82.23 & 80.62 & 80.12 & 79.03 \\
    SigLIP~2 & \underline{86.12} & 86.79 & \underline{87.90} & 85.51 & 85.98 & 86.60 \\
    \rowcolor{oursrow}
    \textbf{\method{} (ours)} & \textbf{86.86} & \textbf{87.25} & \textbf{88.93} & \textbf{86.69} & \textbf{86.34} & \textbf{87.01} \\
    \bottomrule
  \end{tabularx}
  \caption{\textbf{Pooling ablation on UCF-Crime with C-RADIOv4.} Frame AUC of each head when its input is the mean of the full-frame tokens (frame), the mean of the tokens inside $R_t$ (mean) or attention over the patches of $R_t$ (attn.), for each projection alone and for the four projections of \method. Bold/underline: best/second-best per column; ties share the mark.}
  \label{tab:poolabl}
  \vspace{-14pt}
\end{table}

\begin{figure*}[t]
  \centering
  \includegraphics[width=\textwidth]{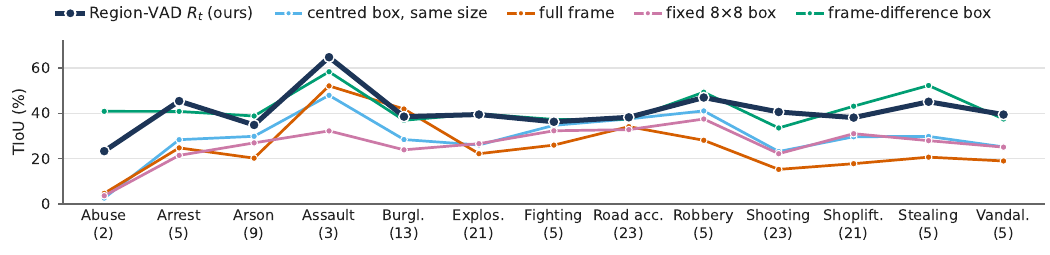}
  \caption{\textbf{Per-class localization on UCF-Crime.} TIoU of $R_t$ and of the four controls of Table~\ref{tab:main} for each anomaly class (number of annotated test videos in parentheses) on the dense boxes of~\cite{abdulaziz2026sst}; Section~B of the appendix repeats it on the boxes of~\cite{liu2019background}. $R_t$ exceeds the three box controls in every class but Burglary, whose annotations cover 42\% of the frame, and the frame-difference box is ahead of $R_t$ in seven classes and behind in six.}
  \label{fig:perclass}
  \vspace{-6pt}   %
\end{figure*}

\begin{figure*}[t]
  \centering
  \includegraphics[width=\textwidth]{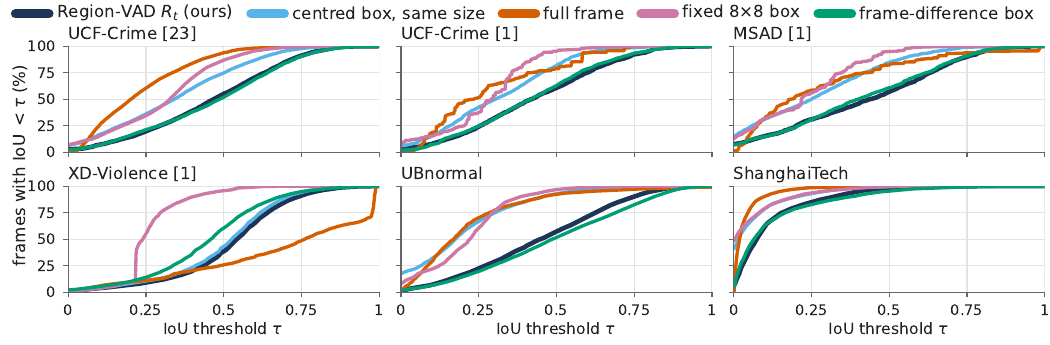}
  \caption{\textbf{Cumulative IoU with the annotated box.} Share of the annotated anomalous test frames whose IoU with the annotated box is below $\tau$ (lower is better), for $R_t$ and the four controls on the annotation sets of Table~\ref{tab:main}. The value at $\tau=0$ is the share of frames on which the box misses the annotated box entirely. The full-frame box always contains the annotated box, so its IoU is the area share of the annotated box and its curve is the distribution of annotated box sizes. Its value at 0 counts the annotated frames beyond the last segment of a video, which no box reaches.}
  \label{fig:miss}
  \vspace{-14pt}
\end{figure*}

\begin{figure*}[t]
  \centering
  \includegraphics[width=\textwidth]{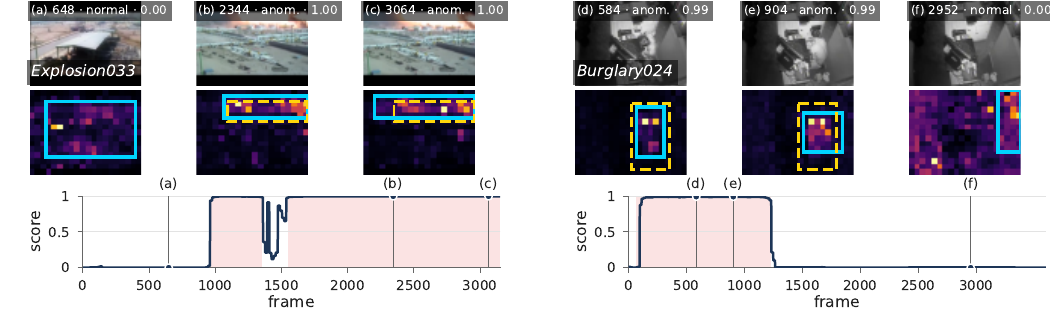}
  \caption{\textbf{Qualitative results of \method on two UCF-Crime test videos.} Top: three frames per video with their index and score. Middle: the change field $\tilde m_t$ of the same segments, with $R_t$ (solid) and the annotation of~\cite{abdulaziz2026sst} (dashed). Bottom: score of the UR-DMU head, with the annotated span shaded and the three frames marked.}
  \label{fig:qual}
  \vspace{-14pt}
\end{figure*}

\keepwithnext{4\baselineskip}
\vspace{-6pt}
\section{Experiments}
\label{sec:exp}
\subsection{Experimental setup}
\label{sec:setup}

\paragraph{Datasets.}
We conduct our experiments on five WS-VAD benchmarks. UCF-Crime~\cite{sultani2018real} and XD-Violence~\cite{wu2020xdv} are the two most widely used real-world datasets, the latter with camera motion. MSAD~\cite{zhu2024msad} is a recent multi-scenario dataset, UBnormal~\cite{acsintoae2022ubnormal} is synthetic with pixel-level masks for every anomaly, and ShanghaiTech~\cite{liu2018future}, in its weakly supervised split~\cite{zhong2019gcn}, is recorded by fixed campus cameras. We evaluate localization on every public spatial annotation, namely the boxes of~\cite{liu2019background} on every 16th UCF-Crime frame, following~\cite{karim2026patch}, the dense per-frame boxes of~\cite{abdulaziz2026sst} on UCF-Crime, MSAD and XD-Violence, and the boxes of the official pixel masks of UBnormal and ShanghaiTech. Table~\ref{tab:boxsize} reports the size of the annotated boxes and of ours; Section~B of the appendix gives the full statistics of the annotated boxes.

\paragraph{Evaluation metrics.}
Detection is measured by frame-level AUC, and frame-level AP on XD-Violence, following the established protocol~\cite{wu2020xdv}, with the score of a segment assigned to its 16 frames. Localization is measured by TIoU, the IoU between the predicted and the annotated box averaged over all annotated anomalous test frames, as defined in~\cite{karim2026patch} and reported as MIoU in~\cite{abdulaziz2026sst}. TIoU ignores the detection score of these frames and does not count false alarms on normal video, which AUC and AP measure.

\paragraph{Controls.}
Since TIoU credits a box both for where it is placed and for how large it is, every localization result of \method is reported next to three controls. The \emph{centred box} is a rectangle of the same size as $R_t$ placed at the centre of the frame, so a gain over it comes from placement. The TIoU of the \emph{full-frame box} is the area fraction of the annotated boxes, and a \emph{fixed box} of $8{\times}8$ patches at the centre of the frame depends on no method. On XD-Violence the annotated boxes are large and central (Table~\ref{tab:boxsize}) and the full frame alone is the second-best localizer of Table~\ref{tab:main}. The fixed box still scores above several published localizers on the boxes of~\cite{liu2019background} and above two on ShanghaiTech. A fourth control, the frame-difference box, applies the same region selection to pixel frame differencing, averaged over the 16 frames of the segment and pooled to the same $20\times15$ grid, to measure what the encoder's tokens add. Every detection result is compared with the \emph{full-frame feature}, the same head trained on the mean of the backbone tokens of the whole frame as in the conventional order of Fig.~\ref{fig:teaser}a, and with the same head trained on the mean of the tokens inside each of the three other boxes. For these boxes, we report the best of the four projections on each dataset.

\paragraph{Implementation details.}
The pooling stages and the head are trained jointly with AdamW for 1,000 steps on batches of four anomalous and four normal videos. Every detection number is the mean over three training seeds, and Section~A of the appendix gives the complete settings and the spread over seeds. On one A100, a 16-frame segment takes 14.9 ms end to end.

\subsection{State-of-the-art comparison}
\label{sec:results}

\paragraph{Localization.}
We compare \method with every published localizer on the five benchmarks in Table~\ref{tab:main}. \method sets a new state of the art on UCF-Crime and MSAD. On the boxes of~\cite{liu2019background} it surpasses the previous best, Patch-MIL~\cite{karim2026patch}, by 5.20 TIoU. On the dense annotations of~\cite{abdulaziz2026sst} it is 12.55 and 6.89 TIoU above the trained localizer of SST-WSVADL~\cite{abdulaziz2026sst} on UCF-Crime and MSAD. On UBnormal, we report the first result on the official masks. These margins hold against the three box controls (last rows of Table~\ref{tab:main}). On UCF-Crime, MSAD and UBnormal, the size-matched centred box and the fixed box score far lower, so the gain comes from where $R_t$ is placed, which supports the first hypothesis of Sec.~\ref{sec:intro}. With pixel differences instead of tokens, the same rule localizes above every trained localizer on every dataset except XD-Violence, so the rule accounts for most of the localization. The token change is ahead on the dense boxes of UCF-Crime and MSAD and under the camera motion of XD-Violence, where frame differencing falls below the centred box, while frame differencing is ahead on the fixed cameras of UBnormal and ShanghaiTech and on the sparse frames of~\cite{liu2019background}. The token change reuses the encoder pass that detection requires. On XD-Violence, where the event fills most of the frame and the camera follows it (Table~\ref{tab:boxsize}), the full-frame box beats our rectangle and every published localizer but one, and \method stays above its centred control. On ShanghaiTech, the annotated events are far smaller than $R_t$ (Table~\ref{tab:boxsize}), which has no size prior, and \method remains 0.6 points below Patch-MIL and above its centred control. Per class (Fig.~\ref{fig:perclass}), $R_t$ beats the three box controls in every UCF-Crime class but Burglary, and the frame-difference box is ahead of $R_t$ in about half of them. In Fig.~\ref{fig:miss}, $R_t$ misses the annotated box entirely on 1.1--7.8\% of the annotated frames. On every set but XD-Violence, it has a lower share of frames below 0.3 and 0.6 IoU than any of the three box controls, and the frame-difference box misses the annotated box less often than $R_t$.

\paragraph{Detection.}
In detection, \method is competitive with the localizing methods of Table~\ref{tab:main}, whose detectors score the full frame. On UCF-Crime its AUC is the highest of the table, above SST-WSVADL~\cite{abdulaziz2026sst} and STPrompt~\cite{wu2024stprompt}. It is below SST-WSVADL with VideoMAE~v2 on MSAD and below VadCLIP~\cite{wu2024vadclip} and both SST-WSVADL models on XD-Violence. These methods also differ from \method in backbone and head, so the effect of reading the full frame is measured with the controls. Against the full-frame feature of the same encoder and head, the region improves detection on all five benchmarks (row $\Delta$ Full frame of Table~\ref{tab:main}), although on ShanghaiTech the gain is within the spread over seeds (Table~S1 of the appendix). The size-matched centred box stays below the region on all five benchmarks, with the learned pooling as well as with the mean of the same rectangle, so the detection gain also comes from where the box is placed. The fixed box and the frame-difference box are also below the region on all datasets with the learned pooling. This supports the second hypothesis of Sec.~\ref{sec:intro}, that the tokens inside the rectangle carry the evidence needed to detect the anomaly.
\begin{table}[t]
  \centering
  \footnotesize
  \renewcommand{\arraystretch}{0.88}
  \setlength{\tabcolsep}{2pt}
  \begin{tabularx}{\columnwidth}{@{}l*{4}{>{\centering\arraybackslash}X}@{}}
    \toprule
    & \multicolumn{2}{c}{TIoU} & \multicolumn{2}{c}{UR-DMU~\cite{zhou2023urdmu}} \\
    \cmidrule(lr){2-3}\cmidrule(lr){4-5}
    Backbone & \tiouLM{} & \tiouSST{} & $R_t$ & frame \\
    \midrule
    \rowcolor{oursrow}
    \textbf{\method{} (ours)} & \textbf{45.71} & \textbf{42.45} & \textbf{88.93} & \underline{85.94} \\
    \quad backbone map & 42.75 & 40.71 & 86.81 & \underline{85.94} \\
    \quad DINOv3 map & \underline{45.60} & \textbf{42.45} & 85.47 & 84.31 \\
    \quad SAM~3 map & 43.26 & 40.36 & 82.23 & 81.99 \\
    \quad SigLIP~2 map & 45.28 & \underline{42.34} & \underline{87.90} & \textbf{86.12} \\
    \midrule
    V-JEPA~2.1 & 42.61 & 41.15 & 84.58 & 81.52 \\
    VideoMAE~v2 & 37.72 & 38.33 & 82.53 & 83.87 \\
    InternVideo2 & 34.56 & 34.48 & 86.92 & 84.91 \\
    SAM~2 & 45.53 & 42.01 & 81.67 & 78.56 \\
    RADIO~v2.5 & 42.15 & 40.23 & 84.64 & 83.39 \\
    \bottomrule
  \end{tabularx}
  \caption{\textbf{Backbone ablation on UCF-Crime.} TIoU of the region selected by the rule of Sec.~\ref{sec:select} from each token set, and frame AUC of UR-DMU with attention pooling inside $R_t$ ($R_t$) or the mean of the full-frame tokens (frame), which for \method is the full-frame control of Table~\ref{tab:main} on the backbone tokens. Indented rows select the region from one map of C-RADIOv4 alone, as in Table~\ref{tab:stage1}, and their detection pools that map's tokens inside the $R_t$ of \method, as in Table~\ref{tab:poolabl}. The other encoders run the whole pipeline on their single change map. Bold/underline: best/second-best per column; ties share the mark.}
  \label{tab:backbones}
\end{table}
\paragraph{Qualitative results.}
Figure~\ref{fig:qual} shows both outputs of \method on two UCF-Crime test videos. The score rises with the event and, in Burglary024, falls after it, and $R_t$ follows the event while the score is high.

\subsection{Ablation studies}
\label{sec:ablation}

\paragraph{Region selection.}
As shown in Table~\ref{tab:stage1}, the temporal smoothing adds 1.5 and 2.3 TIoU on the two annotation sets. The consensus of the four maps matches the best single map, DINOv3, and is behind frame differencing on the boxes of~\cite{liu2019background}, UBnormal and ShanghaiTech (Table~\ref{tab:main}). On UCF-Crime the consensus brings no gain in accuracy, but it combines the four maps instead of selecting one, a choice that would require box annotations.

\paragraph{Pooling.}
Table~\ref{tab:poolabl} ablates the two pooling stages on UCF-Crime with C-RADIOv4. With UR-DMU, mean pooling inside $R_t$ improves on the full-frame mean for every projection, by up to 2.5 AUC, and attention over the patches adds to the mean for the backbone and SigLIP~2 projections. The attention over the four maps of \method gains most from the learned pooling, 1.7 AUC over the mean of the same region. It exceeds the best single map, SigLIP~2, by 1.0 AUC, less than the spread of \method over seeds (Table~S1 of the appendix), and weights the four maps instead of selecting one. Under the ranking loss of the MLP head the pooling matters less, and the region improves on the full-frame mean for only two of the four projections.

\paragraph{Backbones.}
The localization transfers across backbones (Table~\ref{tab:backbones}), from each map of C-RADIOv4 alone to five further frozen encoders, each on its own change map. On both annotation sets V-JEPA~2.1, SAM~2 and RADIO~v2.5 localize above the previous state of the art, Patch-MIL~\cite{karim2026patch} and SST-WSVADL~\cite{abdulaziz2026sst} (Table~\ref{tab:main}), and VideoMAE~v2 and InternVideo2 do so on the dense boxes. The consensus of C-RADIOv4 matches its DINOv3 map in TIoU and gives the highest AUC. Section~A of the appendix gives the token grid of each encoder.

\vspace{-3pt}
\section{Conclusion}
\label{sec:conclusion}
We presented \method, a localize-then-detect approach to WS-VAD. A training-free rule selects the rectangle where the tokens of a frozen encoder change most, and a learned pooling and a standard temporal head detect from the tokens inside it. \method sets a new state of the art in localization on UCF-Crime and MSAD and detects more accurately from inside the rectangle than from the full frame on all five benchmarks. Future work includes a single detector for the five benchmarks and a size prior that transfers across datasets, for events that fill or barely occupy the frame.

\paragraph{Limitations.} The rule assumes that the event is where the frozen tokens change most. Dominant normal motion, such as traffic, can pull the rectangle away from an event elsewhere. A stationary or slow anomaly changes the tokens little, and one rectangle cannot enclose two distant events without the space between them. One middle frame per 16-frame segment can miss a brief event and does not represent motion inside a segment. In anomalous training videos a misplaced rectangle gives the pooling noisy input. Context outside the rectangle, such as signs or lane markings, reaches the detector only through the frozen tokens inside it.

{
    \small
    \bibliographystyle{ieeenat_fullname}
    \bibliography{main}
}

\clearpage
\appendix
\setcounter{table}{0}
\setcounter{figure}{0}
\renewcommand{\thetable}{S\arabic{table}}
\renewcommand{\thefigure}{S\arabic{figure}}
\renewcommand{\theHtable}{S\arabic{table}}
\renewcommand{\theHfigure}{S\arabic{figure}}

\section{Implementation details}
\label{sec:supp-impl}

\paragraph{Encoder and tokens.}
Videos are divided into non-overlapping segments of 16 frames. The middle frame of each segment is resized to $320\times240$ pixels and encoded once by the frozen C-RADIOv4-SO400M~\cite{radiov4} (Sec.~\ref{sec:prelim}), with a patch size of 16, which yields a $20\times15$ grid of tokens for the backbone and for each of its DINOv3, SAM~3 and SigLIP~2 projections. The region $R_t$ is computed once from these tokens with the rule of Sec.~\ref{sec:select} and stored as a set of patches; it is mapped to pixels through the patch grid and reported for every frame of its segment. No part of the localizer is trained or tuned per dataset.

\paragraph{Pooling and head.}
Each projection is brought to $d=512$ dimensions by its own layer normalization and linear layer. Both attention probes of Sec.~\ref{sec:detect} use $H=8$ heads, a single learned query and an MLP with one hidden layer of width $4d$ and GELU activation. The pooled feature is fed to the head's own input layer, a linear layer followed by layer normalization. UR-DMU~\cite{zhou2023urdmu} keeps its published architecture, losses and loss weights. The ablation of Table~\ref{tab:poolabl} also uses the multilayer perceptron of~\cite{sultani2018real} as a lighter head on the $\ell_2$-normalized pooled feature. It is trained with its ranking hinge between the highest score of an anomalous and of a normal video and with its temporal smoothness and sparsity terms at their published weights.

\paragraph{Optimization.}
The pooling stages and the head are trained jointly with AdamW (learning rate $10^{-4}$, weight decay $10^{-2}$, warm-up followed by cosine decay) for 1,000 steps on batches of four anomalous and four normal videos, each with all its segments, without early stopping or checkpoint selection. 

\paragraph{Seeds.}
Every detection number is the mean over three training seeds. Table~\ref{tab:supp-seeds} gives the standard deviation over the seeds for the detection rows of Table~\ref{tab:main}. 

\begin{table*}[t]
  \centering
  \footnotesize
  \setlength{\tabcolsep}{4pt}
  \begin{tabularx}{\textwidth}{@{}l*{5}{>{\centering\arraybackslash}X}@{}}
    \toprule
    & UCF-Crime AUC & MSAD AUC & XD-Violence AP & UBnormal AUC & ShanghaiTech AUC \\
    \midrule
    Full frame & 85.94 $\pm$ 0.15 & 87.98 $\pm$ 0.91 & 71.44 $\pm$ 3.17 & 63.52 $\pm$ 1.27 & 96.96 $\pm$ 0.15 \\
    Centred box & 84.84 $\pm$ 0.50 & 89.03 $\pm$ 0.18 & 75.03 $\pm$ 2.15 & 59.88 $\pm$ 1.43 & 95.61 $\pm$ 0.92 \\
    Fixed $8{\times}8$-patch box & 85.48 $\pm$ 0.32 & 88.39 $\pm$ 0.82 & 71.70 $\pm$ 2.88 & 64.01 $\pm$ 1.71 & 96.55 $\pm$ 0.21 \\
    Frame-difference box & 85.65 $\pm$ 0.22 & 89.55 $\pm$ 0.97 & 76.83 $\pm$ 1.23 & 67.45 $\pm$ 1.41 & 94.90 $\pm$ 0.22 \\
    \method, mean pooling & 87.25 $\pm$ 0.43 & 90.80 $\pm$ 1.22 & 79.73 $\pm$ 2.36 & 67.83 $\pm$ 1.12 & 95.92 $\pm$ 0.65 \\
    \rowcolor{oursrow}
    \textbf{\method (ours)} & \textbf{88.93} $\pm$ 1.21 & \textbf{91.39} $\pm$ 0.92 & \textbf{81.50} $\pm$ 1.41 & \textbf{68.17} $\pm$ 3.18 & \textbf{97.13} $\pm$ 0.61 \\
    \bottomrule
  \end{tabularx}
  \caption{\textbf{Spread over seeds.} Mean and standard deviation over three training seeds of the detection rows of Table~\ref{tab:main}. Bold: best per column.}
  \label{tab:supp-seeds}
\end{table*}

\paragraph{Other encoders.}
For the encoders of Table~\ref{tab:backbones} other than C-RADIOv4, the frame is resized to the encoder's own input size and the rule of Sec.~\ref{sec:select} is applied, without the per-patch minimum, to the single change map of its token grid, which has $24\times24$ patches for V-JEPA~2.1, SAM~2 and RADIO~v2.5, $16\times16$ for InternVideo2 and $14\times14$ for VideoMAE~v2. Pooling, head and training schedule are unchanged.

\paragraph{Motion instead of token change.}
In Table~\ref{tab:motion-boxes}, we replace the token change of Sec.~\ref{sec:select} by motion on UCF-Crime and keep the rest of the rule unchanged. Pixel frame differencing, the frame-difference control of Table~\ref{tab:main}, localizes on par with \method, while the magnitude of RAFT optical flow~\cite{teed2020raft} is about 20 TIoU lower, with or without the median flow removed.

\begin{table*}[t]
  \centering
  \footnotesize
  \setlength{\tabcolsep}{6pt}
  \begin{tabularx}{\textwidth}{@{}l*{7}{>{\raggedleft\arraybackslash}X}@{}}
    \toprule
    & & & \multicolumn{3}{c}{box area, \% of the frame} & \multicolumn{2}{c}{share of frames} \\
    \cmidrule(lr){4-6}\cmidrule(lr){7-8}
    Annotation set & videos & frames & mean & median & IQR & $>$50\% & $<$5\% \\
    \midrule
    UCF-Crime, boxes of~\cite{liu2019background} (every 16th frame) & 140 & 3,102 & 23.1 & 20.1 & 9.9--33.5 & 6\% & 8\% \\
    UCF-Crime, dense boxes of~\cite{abdulaziz2026sst} & 140 & 82,584 & 28.4 & 23.5 & 12.1--41.2 & 21\% & 3\% \\
    MSAD, dense boxes of~\cite{abdulaziz2026sst} & 120 & 31,278 & 28.0 & 19.4 & 7.7--44.4 & 18\% & 15\% \\
    XD-Violence, dense boxes of~\cite{abdulaziz2026sst} & 500 & 531,398 & 69.1 & 76.2 & 49.1--97.8 & 74\% & 3\% \\
    UBnormal, official masks & 158 & 68,098 & 21.8 & 17.1 & 8.8--28.7 & 8\% & 13\% \\
    ShanghaiTech, official masks & 44 & 7,962 & 3.9 & 1.9 & 0.7--4.6 & 0\% & 77\% \\
    \bottomrule
  \end{tabularx}
  \caption{\textbf{Size of the annotated boxes.} Area of the ground-truth box as a percentage of the frame over the annotated anomalous test frames: mean, median, interquartile range, and the share of frames whose box exceeds half the frame or is below 5\% of it.}
  \label{tab:supp-boxsize}
\end{table*}

\begin{table}[t]
  \centering
  \footnotesize
  \setlength{\tabcolsep}{2pt}
  \begin{tabularx}{\columnwidth}{@{}l*{3}{>{\centering\arraybackslash}X}@{}}
    \toprule
    & \multicolumn{2}{c}{UCF-Crime} & \\
    \cmidrule(lr){2-3}
    Box source & \tiouLM{} & \tiouSST{} & FPS \\
    \midrule
    frame-difference box & \textbf{45.96} & \underline{41.25} & \textbf{721} \\
    optical flow~\cite{teed2020raft} & 25.14 & 22.77 & 61 \\
    optical flow, median flow removed & 24.40 & 22.01 & 61 \\
    \midrule
    \rowcolor{oursrow}
    \textbf{\method} & \underline{45.71} & \textbf{42.45} & \underline{715} \\
    \bottomrule
  \end{tabularx}
  \caption{\textbf{Region selection from motion and from tokens on UCF-Crime.} TIoU of the rectangle selected by the rule of Sec.~\ref{sec:select}, with the same temporal smoothing, from pixel frame differencing, from RAFT-large optical-flow magnitude over the 16 frame pairs of each segment, each frame with its predecessor (raw, and with the per-pair median flow removed at a cost negligible next to RAFT), and from the consensus of the four projections (\method). FPS counts video frames per second of the whole pipeline, encoder and detection head included, on one NVIDIA L4 with segments processed one at a time and video decoding excluded. On an A100, \method runs at 1,076 FPS (Sec.~\ref{sec:setup}). Bold/underline: best/second-best per column.}
  \label{tab:motion-boxes}
\end{table}
\section{Size of the annotated boxes}
\label{sec:supp-boxsize}
\begin{figure*}[t]
  \centering
  \includegraphics[width=\textwidth]{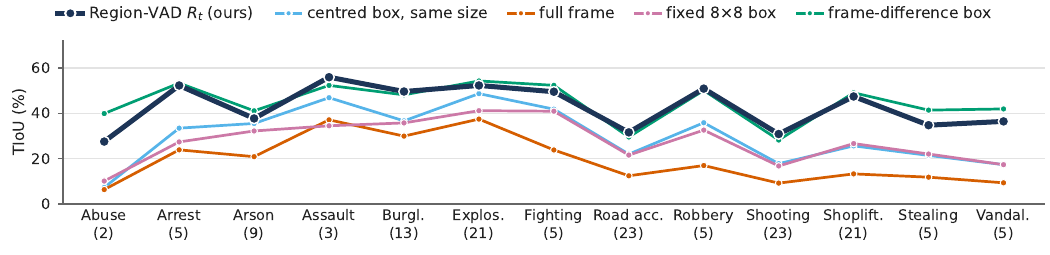}
  \caption{\textbf{Per-class localization on UCF-Crime on the boxes of~\cite{liu2019background}.} Counterpart of Fig.~\ref{fig:perclass} on every 16th frame. TIoU of $R_t$ and of the four controls of Table~\ref{tab:main} for each anomaly class (number of annotated test videos in parentheses). $R_t$ exceeds the three box controls in every class, and the frame-difference box is ahead of $R_t$ in eight classes and behind in five.}
  \label{fig:supp-perclass}
\end{figure*}

Table~\ref{tab:supp-boxsize} gives the size of the ground-truth boxes of every annotation set of Sec.~\ref{sec:setup}, as a fraction of the frame, and completes Table~\ref{tab:boxsize} with the interquartile range and the two tails. A box covering the full frame obtains a TIoU close to the mean box area (Table~\ref{tab:main}). On XD-Violence three quarters of the annotated frames have a box larger than half the frame, so the full-frame box beats $R_t$ and every published localizer but one (Sec.~\ref{sec:results}). ShanghaiTech is the opposite case, with a median box below 2\% of the frame. Figure~\ref{fig:supp-perclass} repeats the per-class analysis of Fig.~\ref{fig:perclass} on the boxes of~\cite{liu2019background}.

\end{document}